\documentclass[journal,10pt]{IEEEtran}

\usepackage{framed,multirow,booktabs}
\usepackage{graphicx,epsfig}
\usepackage{amssymb,amsmath,bm}
\usepackage{textcomp}
\usepackage{amsthm}
\usepackage{lineno,hyperref}

\usepackage{url}
\usepackage{xcolor}
\definecolor{newcolor}{rgb}{.8,.349,.1}

\ifCLASSINFOpdf
\else
\fi

\begin{document}

\title{Multimodal Pre-training Based on Graph Attention Network for Document Understanding}
%
%
%

\author{Zhenrong~Zhang,
	Jiefeng~Ma, Jun~Du, Licheng~Wang and Jianshu~Zhang
	\thanks{Zhenrong Zhang, Jiefeng Ma and Jun Du were with the National Engineering Research Center of Speech and Language Information Processing (NERC-SLIP), University of Science and Technology of China, Hefei, Anhui, China. Jianshu Zhang and Licheng Wang were with the IFLYTEK Research, Hefei, Anhui, China. e-mail: zzr666@mail.ustc.edu.cn, jfma@mail.ustc.edu.cn, jundu@ustc.edu.cn, lcwang2@iflytek.com, jszhang6@iflytek.com. (Corresponding author: Jun Du.)
}}

\maketitle

\begin{abstract}
  Document intelligence as a relatively new research topic supports many business applications. Its main task is to automatically read, understand, and analyze documents. However, due to the diversity of formats (invoices, reports, forms, etc.) and layouts in documents, it is difficult to make machines understand documents. In this paper, we present the GraphDoc, a multimodal graph attention-based model for various document understanding tasks. GraphDoc is pre-trained in a multimodal framework by utilizing text, layout, and image information simultaneously. In a document, a text block relies heavily on its surrounding contexts, accordingly we inject the graph structure into the attention mechanism to form a graph attention layer so that each input node can only attend to its neighborhoods. The input nodes of each graph attention layer are composed of textual, visual, and positional features from semantically meaningful regions in a document image. We do the multimodal feature fusion of each node by the gate fusion layer. The contextualization between each node is modeled by the graph attention layer. GraphDoc learns a generic representation from only 320k unlabeled documents via the Masked Sentence Modeling task. Extensive experimental results on the publicly available datasets show that GraphDoc achieves state-of-the-art performance, which demonstrates the effectiveness of our proposed method. The code is available at \url{https://github.com/ZZR8066/GraphDoc}.
\end{abstract}


%
\IEEEpeerreviewmaketitle

\section{Introduction}
	\IEEEPARstart{A}{s} an indispensable research area in NLP, document understanding aims to automate the information extraction from documents and support numerous business applications. This technology can significantly reduce the laborious document process workflows through automated document classification, entity recognition, semantic extraction, etc.

\begin{figure*}[t]
		\centerline{\includegraphics[width=1\linewidth]{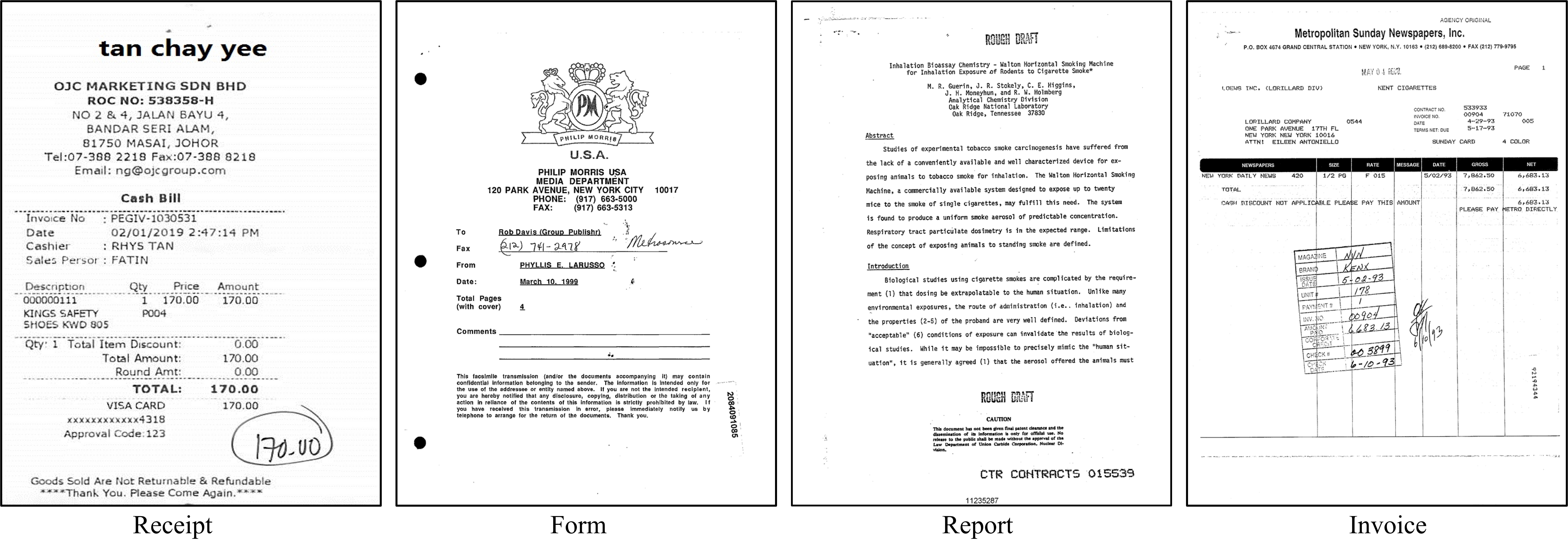}}
		\caption{The example images of documents with diﬀerent formats and layouts. The layout and visual content of different documents are markedly inconsistent.}
		\label{introduction}
	\end{figure*}

	Documents convey information through plain text, visual content, and layout structure. As shown in Figure~\ref{introduction}, documents include a variety of types such as receipts, forms, invoices, and reports. Different types of documents indicate that the text fields of interest are located at different positions within the document, which is often determined by the style and format of each type as well as the document content. Therefore, to precisely understand documents, it is inevitable to take advantage of the cross-modality nature of documents, where the textual, visual, and layout information should be jointly modeled and learned in a multimodal framework~\cite{01multimodal,02multimodal,03multimodal}. 
	
	Self-supervised learning has emerged as a paradigm to learn general data representations from unlabeled examples and to fine-tune the model on labeled data~\cite{01selfsupervised,02selfsupervised,03selfsupervised}. This has been verified successfully in a variety of NLP tasks~\cite{bert,roberta} in recent years. Despite the widespread use of pre-training models for NLP applications, they focus almost exclusively on text-level manipulation, while neglecting image and layout that is vital for document understanding. Recently, many pre-training models~\cite{layoutlm,layoutlmv2,selfdoc,unidoc} modified the BERT~\cite{bert} architecture by combining textual features with images and layouts. These approaches achieved state-of-the-art results in several document understanding tasks~\cite{funsd,cord,rvlcdip}, which demonstrate the effectiveness of multimodal self-supervised pre-training. Additionally, from a practical perspective, many tasks related to document understanding are label-scarce. Therefore, applying the self-supervised pre-training to learn a generic representation from a collection of unlabeled documents in a multimodal framework is essential.

	Most contemporary BERT-like pre-training models for document understanding~\cite{layoutlm,layoutlmv2,docformer,bros} use individual words as inputs. In a document, however, a single word can be understood within the local contexts and does not always require analyzing the entire page. With all words in a document considered, these models will not be sufficiently penalized during the pre-training phase. Moreover, these pre-training models will suffer from input length constraints, especially for text-rich documents. In our work, we follow Self-Doc~\cite{selfdoc} and deem semantic regions (text block, table, heading, etc.) in document images as basic input elements instead of words.

	Although self-attention~\cite{transformer} is a basic yet powerful component in the Transformer architecture, it is inefficient to some extent. As each input element has to attend to all $n$ elements, the overall complexity scales as $\mathcal{O}(n^2)$. In a document, however, a semantic region relies more heavily on its surrounding context, which is already a robust inductive bias. However, previous works~\cite{layoutlmv2,selfdoc,unidoc} apply the Transformer to learn this bias from scratch during the pre-training phase, which increases the learning cost. Therefore, how to leverage this prior knowledge to ``lighten up'' the pre-training model will be meaningful. In our work, we inject the graph structure in a document into the attention mechanism to form the graph attention layer instead of the original Transformer architecture to mitigate this problem.
	
\begin{figure*}[htp]
		\centerline{\includegraphics[width=.9\linewidth]{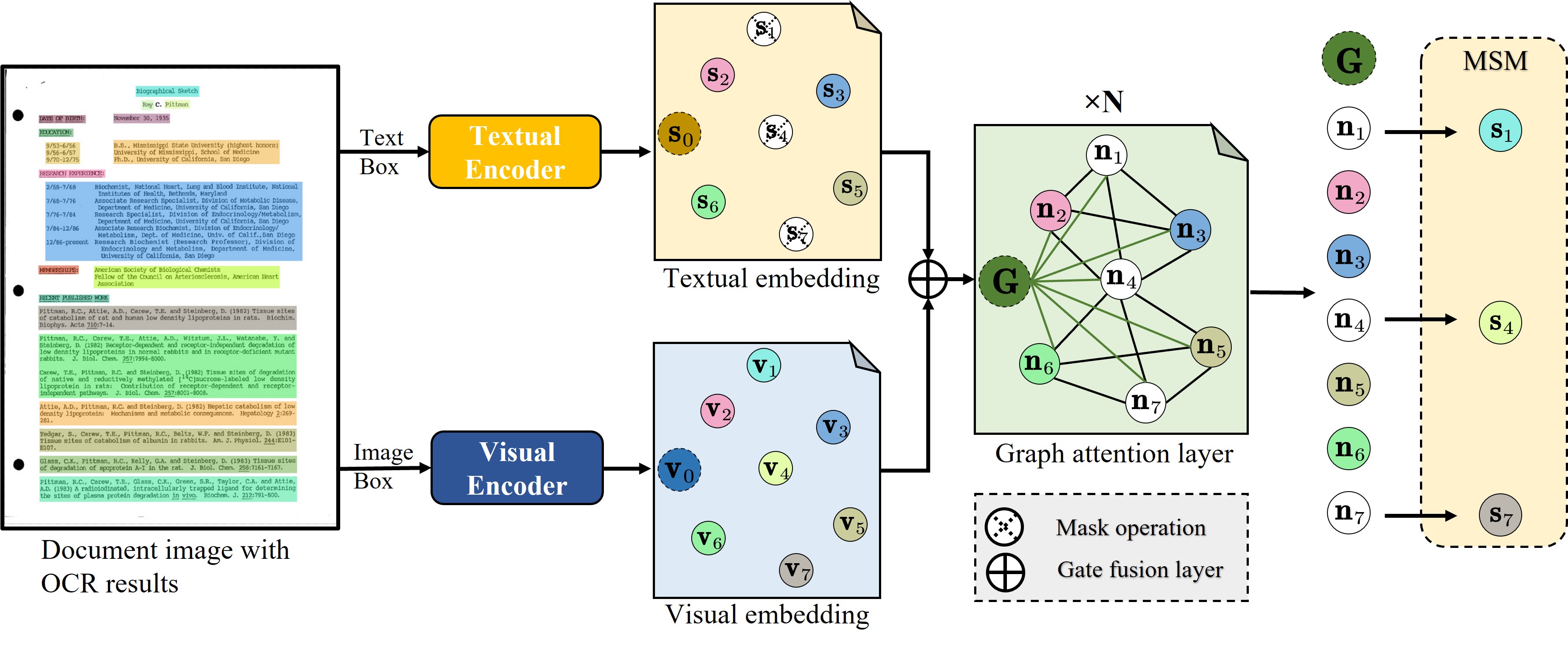}}
		\caption{An illustration of the multimodal framework of GraphDoc. Given a document image with OCR results, we first extract the corresponding textual and visual embedding for each text region using the textual encoder and the visual encoder, respectively. Then, we apply the well-designed graph attention layer to encode the multimodal representations for each region. The model is pre-trained via the Masked Sentence Modeling (MSM) task.}
		\label{overview}
	\end{figure*}

	In this paper, we present the GraphDoc, a multimodal graph attention-based model for document understanding as shown in Figure~\ref{overview}. GraphDoc follows the now common, pre-training and fine-tuning strategy. We treat semantic regions of document images extracted by the Optical Character Recognition (OCR) as basic input elements instead of words. Distinct from previous pre-training models~\cite{layoutlm,bros,structurallm}, which only focus on combining textual features with corresponding layouts, we fully exploit text, image, and layout information during the pre-training phase to learn the cross-modality interaction. More specifically, for each semantic region, we extract textual features using the pre-trained Sentence-BERT~\cite{sentencebert} and apply the RoIAlign~\cite{maskrcnn} to extract the visual features from the output of the visual backbone~\cite{swintransformer}. The final sentence embeddings and visual embeddings are obtained by combining textual features and visual features with spatial layout features, respectively. Different from previous works~\cite{layoutlm,layoutlmv2,docformer}, we design the graph attention network with the gate fusion layer to do multimodal interaction instead of the Transformer architecture. We first do multimodal fusion through the designed gate fusion layer to fuse the sentence embeddings and visual embeddings. Moreover, we make visual information accessible across graph attention layers which act as a residual connection~\cite{resnet}. Then, each input node, which contains both text and image information, does the attention mechanism only on its neighborhoods through the graph attention layer. In addition, the global node, which is attended to each input node, will assist the model to understand documents in a global aspect. As for the pre-training strategy, we simply use the Masked Sentence Modeling (MSM) task. In this way, GraphDoc learns a generic multimodal representation only from 320k unlabeled documents images. 
	
	The main contributions of this paper are as follows:
	\begin{itemize}
		\item We present a multimodal graph attention-based model, named GraphDoc, for document understanding. GraphDoc fully exploits the textual, visual, and positional information of every semantically meaningful region in a document.
		
		\item We inject the graph structure in documents into the attention mechanism to help each input node fully understand documents from both local and global aspects. The ablation studies also demonstrate the effectiveness of the proposed graph attention layer.
		
		\item Extensive experiments show that GraphDoc outperforms other methods by using only 320k document images for pre-training and achieves new state-of-the-art results in some downstream tasks of document understanding.
	\end{itemize}
	
\section{Related works}
	\subsection{Attention mechanism}
	The attention mechanisms as an integral part of models enable neural networks to focus more on relevant elements of the input than on irrelevant parts. In multi-modal tasks, attention mechanism is also widely adopted to capture the cross-modality interaction. \cite{esenet} introduces an expansion-squeeze-excitation (ESE) attention mechanism to aggregate the most discriminative features from RGB and skeleton modalities, for video-based elderly activity recognition. Considering both the spatial and temporal relations of human skeleton motions, \cite{scrnn} proposes a novel skeleton-joint attention with RNNs to achieve better performance in the task of human motion prediction. \cite{huang2016attention} presented a neural machine translation architecture associating visual and textual features for translation tasks with multiple modalities. 
	\cite{nam2017dual}  proposed dual attention networks which jointly leverage visual and textual attention mechanisms to capture fine-grained interaction between vision and language for visual question answering and image-text matching tasks.	\cite{jin2017multimodal} presented a recurrent neural network with an attention mechanism to fuse multimodal features, where image features are incorporated into the joint features of text and social context to produce a reliable fused classification for effective rumor detection.

	While self-attention is powerful, the computation and memory overhead of the Transformer are quadratic to a sequence length. To reduce the complexity in self-attention, some sparse Transformers have been recently proposed. Star-Transformer~\cite{startransformer} replaces the fully-connected structure with a star-shaped topology, in which every two non-adjacent nodes are connected through a shared relay node. Longformer~\cite{longformer} uses a number of efficient attention patterns on the encoder network and reduces the model complexity. Graph attention network~\cite{gat} computes the hidden representations of each node in the
	graph, by attending over its neighbors.
	
	\subsection{Self-supervised learning}
	Recently, self-supervised learning has emerged as an effective technique for settings where labeled data is scarce.  The key idea is to learn general representations in a setup where substantial amounts of unlabeled data are available and to leverage the learned representations to improve performance on a downstream task for which the amount of labeled data is limited. This has been particularly successful for natural language processing~\cite{bert,roberta}, speech recognition~\cite{wav2vec1,wav2vec2} and computer vision~\cite{moco,mocov2}. 
	It is also an active research area for multi-modal tasks such as video action recognition, audio event classification, and text-to-video retrieval. \cite{akbari2021vatt} presented a framework for learning multimodal representations from unlabeled data using convolution-free transformer architectures. \cite{chen2021multimodal} extended the concept of instance-level contrastive learning with a multimodal clustering step in the training pipeline to capture semantic similarities across modalities. 
	\cite{radford2021learning} aimed at learning directly from raw text about images which leverages a much broader source of supervision. It demonstrated that the simple pre-training task of predicting which caption goes with which image is an efficient and scalable way to learn SOTA image representations from scratch.
	
	\subsection{Docuemnt pre-training}
	Document pre-training methods in the literature can be divided into three categories according to the utilization of text, image, and spatial information in document images during the pre-training phase. 
	
	The first is text-based pre-trained models. BERT~\cite{bert}, whose architecture is a multi-layer bidirectional Transformer encoder based on~\cite{transformer}, uses masked language models to obtain pre-trained deep bidirectional representations. The pre-trained BERT model can be finetuned with fewer labeled data and achieve state-of-the-art results for a wide range of NLP tasks.	\cite{roberta} finds that BERT was significantly undertrained and proposes an improved recipe for training BERT models, which is called RoBERTa.
	
	The second is to combine the textual features with the spatial layout. LayoutLM~\cite{layoutlm} is the first to jointly model interactions between text and layout information in a single framework for document-level pre-training. It modifies the BERT architecture by adding 2D spatial coordinate embeddings along with 1D positional and semantic embeddings. BROS~\cite{bros}, which is also a BERT-based encoder, utilizes relative positions between text blocks for spatial layout encoding. It also proposes a novel area-masking self-supervision strategy that reflects the 2D natures of text blocks.	Different from BROS and LayoutLM, StructuralLM~\cite{structurallm} uses cell-level 2D-position embeddings with tokens in a cell sharing the same 2D coordinate. It also proposes a new pre-training object called cell position classification, in addition to the masked visual-language model.
	
	The third is to fully exploit the textual, visual, and positional information of every semantically meaningful component in a document.
	LayoutLMv2~\cite{layoutlmv2} improves over the LayoutLM by integrating the image information with text and layout, and takes advantage of the Transformer architecture to learn the cross-modality interaction between visual and textual information during the pre-training stage. 
	Due to spatial and visual dependencies that might differ across transformer layers, DocFormer~\cite{docformer} unties visual, text, and spatial features. Distinct from previous methods, Self-Doc~\cite{selfdoc} adopts semantically meaningful components (e.g., text block, heading, figure) as the model input instead of isolated words. It takes the pre-extracted RoI features and sentence embeddings as input, and models the perform learning over the textual and visual information using the cross-modality encoder. UniDoc~\cite{unidoc} improves the Self-Doc by making use of three self-supervised tasks, encouraging the representation to model sentences, learn similarities, and align modalities.
	
	Since the task requires understanding texts in various layouts, the combination of multiple technical components from both computer vision and natural language processing is required. In our work, we combine textual features with their image and layout for each semantic region. The GraphDoc is simply pre-trained on the Masked Sentence Modeling (MSM) task and achieves new state-of-the-art on downstream tasks of document understanding.

\section{Method}
	An overview of the multimodal framework of GraphDoc is presented in Figure~\ref{overview}. Given a document image $I$ with $n$ semantic regions, we apply the off-the-shelf OCR engine~\cite{easyocr} to obtain the $i$-th semantic region with the bounding box $b_i$ and its corresponding text sentence $t_i$. For each semantic region, it contains text and image along with its positional information. 
	We design the textual encoder to encode both text and spatial layout information simultaneously to generate sentence embeddings. Similar to the textual encoder, the visual encoder encodes both image and positional information to generate visual embeddings. Then, we stack $N$ blocks which are composed of gate fusion layers and graph attention layers to generate multimodal contextualized representations for all semantic regions. 
	The feature fusion from each modality is performed by the gate fusion layer, while the graph attention layer captures the contextualization information between each region.
	Considering the phenomenon that a text block relies more heavily on its surrounding contexts, the designed graph attention layer allows each region to attend to only its neighbor area $\mathcal{N}\left( i \right)$. In the pre-training stage, the model is pre-trained via the MSM task on a large collection of document images and the generated representation can be further utilized for downstream document understanding tasks.

	\subsection{Textual encoder}
	
	\begin{figure}[t]
		\centerline{\includegraphics[width=1.\linewidth]{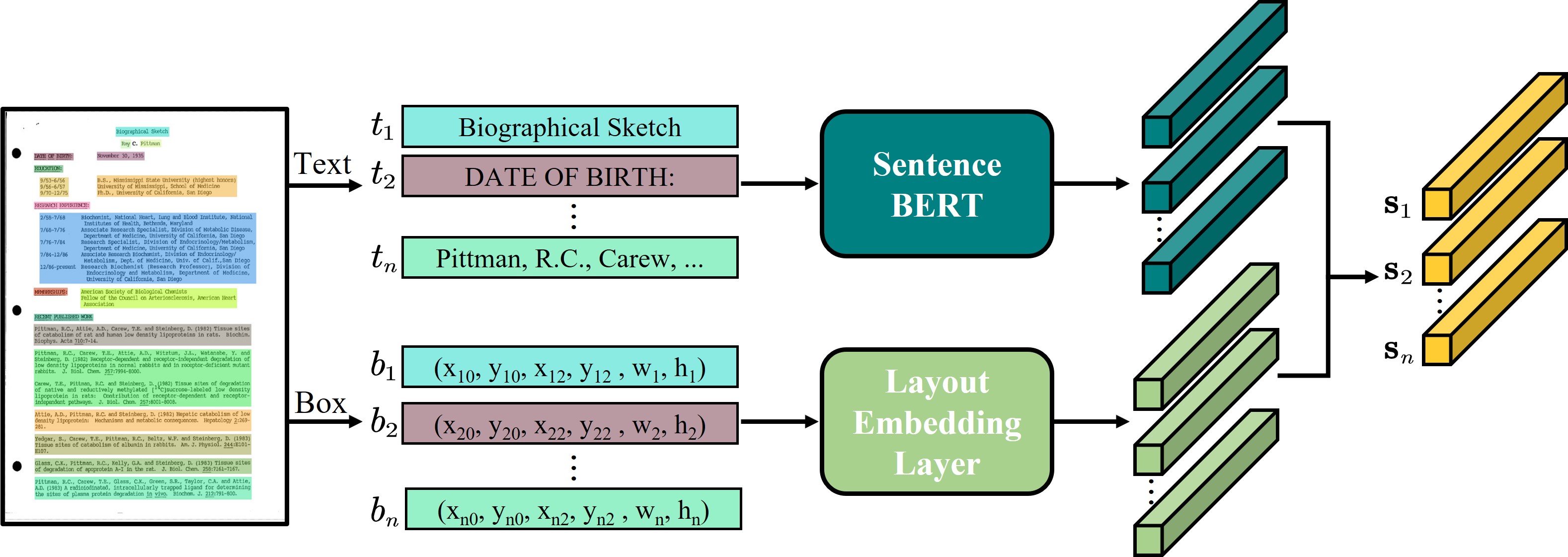}}
		\caption{The illustration of the textual encoder. It applies the Sentence BERT and Layout Embedding Layers to encode both semantic and spatial information, which are then added to form the final sentence embeddings $\mathbf{S}$.}
		\label{textualencoder}
	\end{figure}

	Since the text content in a document is presented in the 2D structure, it is necessary to encode text with layout information. Following the LayoutLMv2~\cite{layoutlmv2}, we normalize and discretize all coordinates to integers in the range of  $[0,512]$, and use two embedding layers to embed x-axis features and y-axis features separately. Given the normalized bounding box of the $i$-th semantic region $b_i$, we calculate the width and height of the box denoted as $w_i$ and $h_i$. The coordinate of four vertices is represented as $(x_{iv}, y_{iv}), v=\{0,1,2,3\}$ in a clockwise manner, starting from the upper left corner. The final 2D layout embedding $\mathbf{l}_i$ is then constructed by concatenating six bounding box features $(x_{i0},y_{i0},x_{i2},y_{i2},w_i,h_i)$ through two layout embedding layers.
	\begin{equation}
		\mathbf{l}_i=\left[ \text{Emb}_x\left( x_{i0}, x_{i2}, w_i \right) ; \text{Emb}_y\left( y_{i0}, y_{i2}, h_i \right) \right], 0\le i\le n
	\end{equation}
	$\left[  ;  \right]$ is the concatenation operation. $\text{Emb}_{\text{x}}$ and $\text{Emb}_{\text{y}}$ are two layout embedding layers. It is worth noting that the corresponding bounding box features for $\mathbf{l}_0$ are $(0,0,\text{W},\text{H},\text{W},\text{H})$, in which $\text{W}$ and $\text{H}$ represent the width and height of the input document image, respectively
	
	As shown in Figure~\ref{textualencoder}, we embed plain text contained in a semantic region into a feature vector using the pre-trained Sentence-BERT model~\cite{sentencebert}, which can derive semantically meaningful sentence embeddings. Parameters of the Sentence-BERT do not update during the pre-training phase. Sentence embeddings $\mathbf{S}$ are calculated as follow:
	\begin{equation}
		\mathbf{s}_i=\text{Proj}\left(\text{SentenceEmb}\left(t_i \right)\right) +\mathbf{l}_i ,0\le i\le n
	\end{equation}
	where SentenceEmb and Proj represent the Sentence-BERT and a linear projection layer, respectively. It is worth noting that $\mathbf{s}_0$ is the [CLS] embedding.

	\subsection{Visual encoder}
	
	\begin{figure}[t]
		\centerline{\includegraphics[width=1.\linewidth]{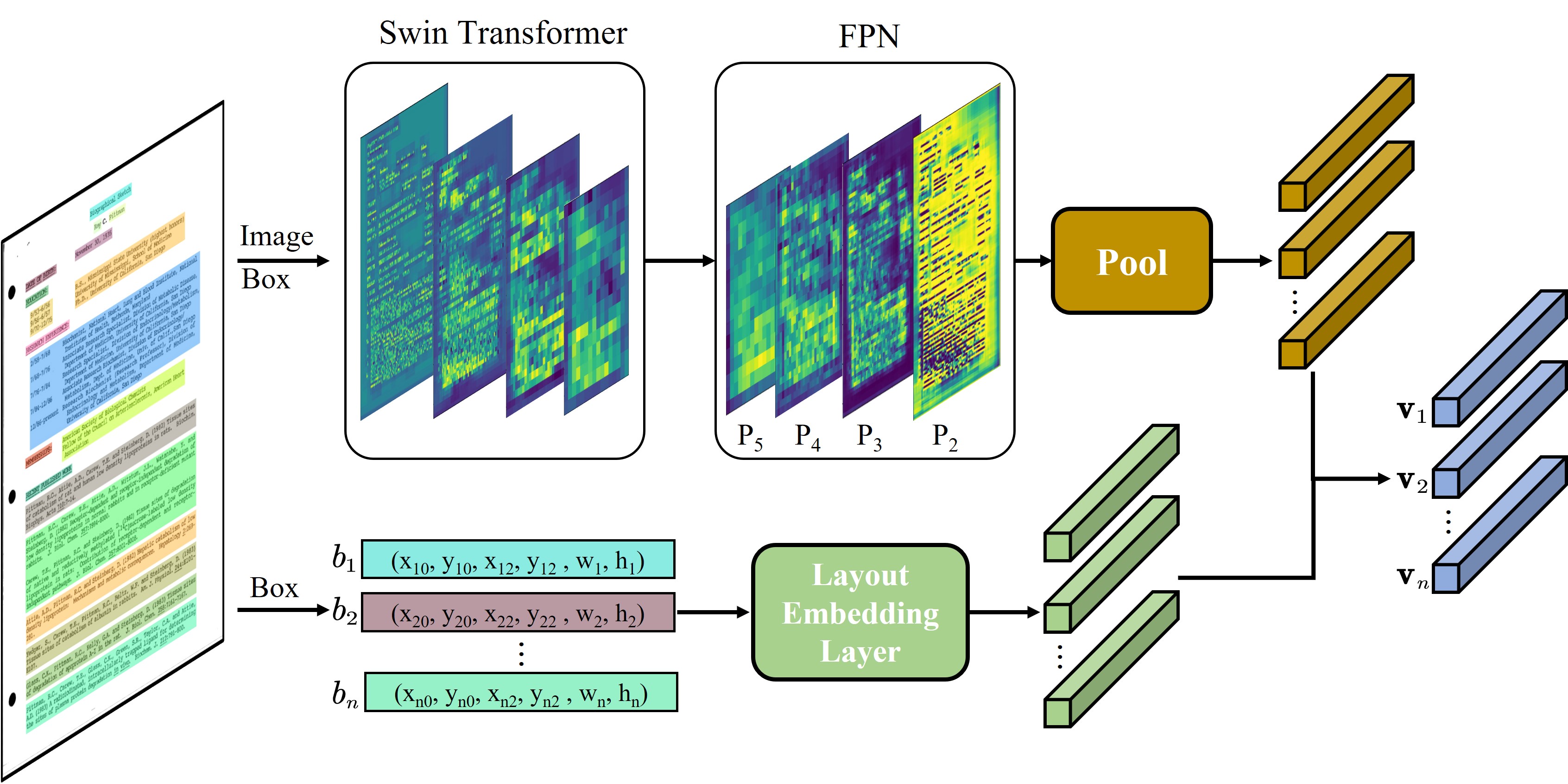}}
		\caption{The illustration of the visual encoder. It applies Swin Transformer with FPN to extract appropriate visual features for the document image. The final visual embeddings are obtained by adding the visual features gained from $\mathbf{P}_2$ with corresponding layout embeddings in each region.}
		\label{visual_encoder}
	\end{figure}

	We use the Swin Transformer~\cite{swintransformer} with FPN~\cite{fpn} as the backbone of our visual encoder. The backbone is first pre-trained on the PubLayNet~\cite{publaynet} dataset to make the extracted visual features more semantics. A document image $I$ is resized to $512\times 512$ then fed into the visual backbone to generate a feature pyramid with four feature maps $\{\mathbf{P}_2, \mathbf{P}_3, \mathbf{P}_4, \mathbf{P}_5\}$ as shown in Figure~\ref{visual_encoder}. The output $\mathbf{P}_2$ is the feature map from FPN with 1/4 size of the input image. After that, the image feature of each semantic region is extracted from $\mathbf{P}_2$ by RoIAlign~\cite{maskrcnn} according to $b_i$. The visual embedding $\mathbf{v}_i$ is computed as follow:
	\begin{equation}
		\mathbf{v}_i=\text{Proj}\left( \text{Pool}\left( \text{Backbone}\left( I \right) ,b_i \right) \right) +\mathbf{l}_i ,0\le i\le n
	\end{equation}
	where Proj is a linear projection layer applied to each region-level image feature in order to unify the dimensions. Pool represents the RoIAlign operation. It is worth noting that $\mathbf{v}_0$ is the average of $\mathbf{P}_2$, which is used to represent the information of the whole image.

	\begin{figure*}[t]
		\centerline{\includegraphics[width=.9\linewidth]{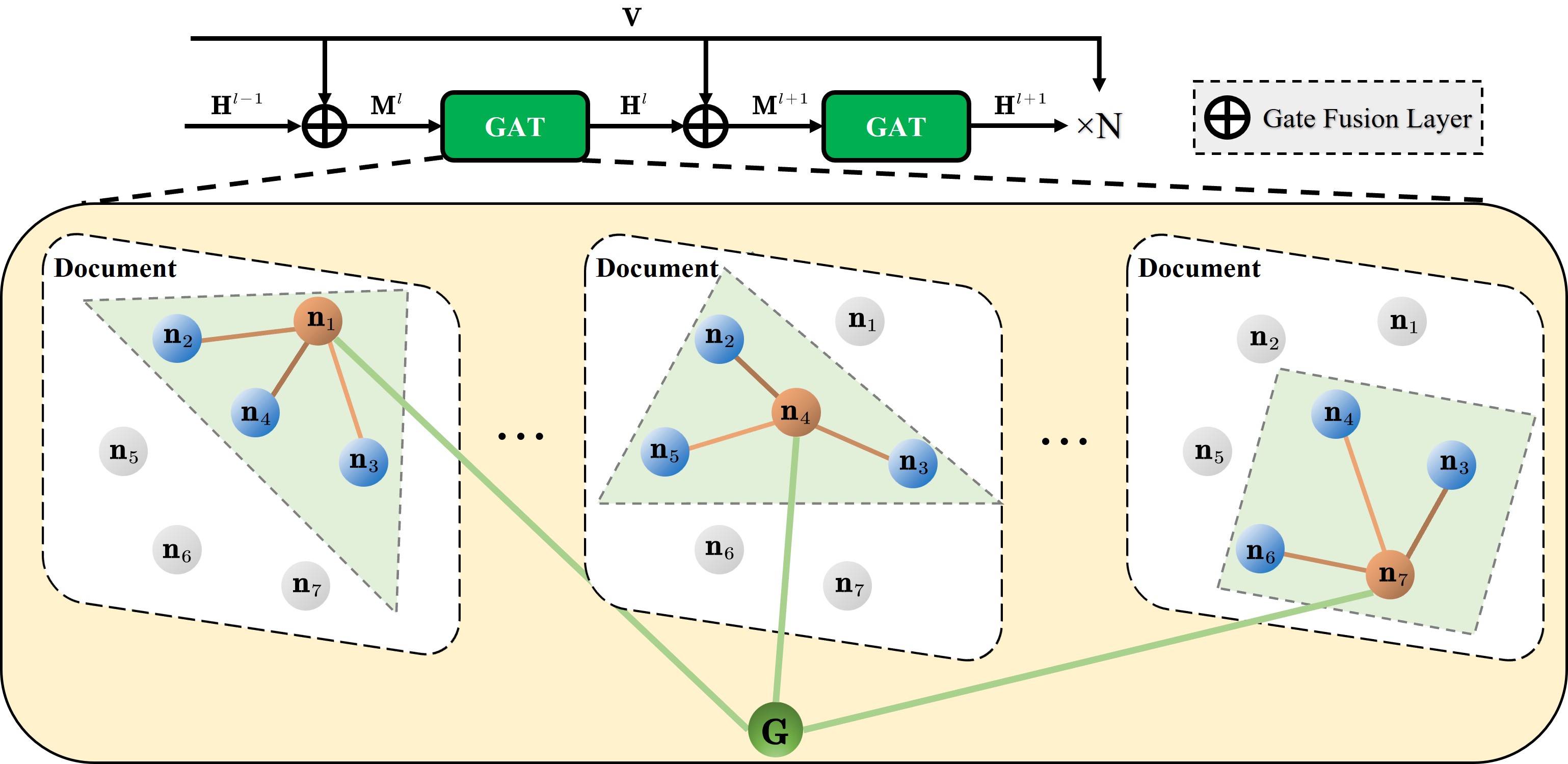}}
		\caption{An illustration of the graph attention layer. The $\{\textbf{n}_1-\textbf{n}_7\}$ represent different input nodes of the layer. \textbf{G} is the global node, which assists each node to capture the global information of the document. Graph attention mechanism employed on $\{\textbf{n}_1,\textbf{n}_4,\textbf{n}_7\}$ is visualized. $\{\textbf{n}_1,\textbf{n}_4,\textbf{n}_7\}$ only attend to their neighborhood nodes and a global node G.}
		\label{gat}
	\end{figure*}

	\subsection{Gate fusion layer}
	Most previous pre-training models~\cite{layoutlmv2,structext} produce an embedding sequence by collecting multimodal information from text, vision, and layout, and then perform a transformer network to establish deep fusion on different modalities. In our work, we adopt semantically meaningful components (e.g., text block, table, figure) as the model input. Since each component has its corresponding multimodal information, we design the gate fusion layer to explicitly fuse information from each modality. 
	
	Moreover, we believe that the dependencies between text and image might differ across graph attention layers, which is verified in our ablation experiments. Inspired by ResNet~\cite{resnet}, we make visual information accessible across graph attention layers act as an information residual connection. The gate fusion layer is designed as follow:
	\begin{align}
		& z_i^{l}=\sigma \left( \mathbf{W}_2 g\left( \mathbf{W}_1\left[\mathbf{v}_i ; \mathbf{h}_{i}^{l-1} \right] +\mathbf{b}_1 \right) +\mathbf{b}_2 \right) \\
		& \mathbf{m}_{i}^{l}=\left( 1-z_i^{l} \right) \mathbf{h}_{i}^{l-1}+z_i^{l}\mathbf{v}_i
	\end{align}
	where $\mathbf{m}_{i}^{l}\in \mathbb{R}^d$, $\mathbf{h}_{i}^{l}\in \mathbb{R}^d$, $\mathbf{W}_1\in \mathbb{R}^{d\times 2d}$, $\mathbf{b}_1\in \mathbb{R}^{d}$, $\mathbf{W}_2\in \mathbb{R}^{1\times d}$, $\mathbf{b}_2\in \mathbb{R}^1$. $d$ is the dimension of the visual embedding. The $\sigma$ and $g$ are sigmoid and GELU~\cite{gelu} function. As shown in the upper of Figure~\ref{gat}, $\mathbf{m}_{i}^{l}$ represents the $i$-th output element in the $l$-th gate fusion layer. $\mathbf{h}_{i}^{l}$ represents the $i$-th output hidden representation in the $l$-th graph attention layer. It is worth noting that $\mathbf{h}_{i}^{0} = \mathbf{s}_{i}$. 

	\subsection{Graph attention layer}
	The observation that a text block in a document relies more heavily on its surrounding context is a robust inductive bias. However, previous pre-trained models~\cite{layoutlm,layoutlmv2,selfdoc,unidoc} apply the Transformer to learn this bias from scratch during the pre-training stage. Inspired by GAN~\cite{gat} and StartTransformer~\cite{startransformer}, we design the graph attention layer to compute the hidden representation of each node in the graph, by attending over its neighbors following a self-attention strategy. As shown in Figure~\ref{gat}, each node attends to only its neighborhood nodes and a global node, which can assist the model to understand the document from both local and global aspects.
	
	The input to $l$-th graph attention layer is features of $n$ nodes, $\mathbf{M}^{l}=\left\{ \mathbf{m}_{1}^{l},\mathbf{m}_{2}^{l},...,\mathbf{m}_{n}^{l} \right\}$. The layer produces a new set of node features $\mathbf{H}^{l}=\left\{ \mathbf{h}_{1}^{l},\mathbf{h}_{2}^{l},...,\mathbf{h}_{n}^{l} \right\}$, as its output. 
	Following the original self-attention mechanism~\cite{transformer}, we calculate the attention score between the $j$-th node and $i$-th node as follows:
	\begin{equation}
		{e}_{ij}=\left( \mathbf{W}^{\text{q}}\mathbf{m}_{i}^{l} \right) ^{\top} \left( \mathbf{W}^{\text{k}}\mathbf{m}_{j}^{l} \right)
	\end{equation}
	where $\mathbf{W}^{\text{q}}\in \mathbb{R}^{d\times d}$, $\mathbf{W}^{\text{k}}\in \mathbb{R}^{d\times d}$. In addition, inspired by TransformerXL~\cite{transformerxl} and BROS~\cite{bros}, we explore a relative position encoding in 2D structure to improve the attention mechanism. The relative position encoding between $i$-th node and $j$-th node is calculated as $\mathbf{p}_{ij}=\left[ \mathbf{f}^{\text{sinu}}\left( x_{iv}-x_{jv} \right) ;\mathbf{f}^{\text{sinu}}\left( y_{iv}-y_{jv} \right) \right]$. 
	Here $\mathbf{f}^{\text{sinu}}$ indicates a sinusoidal function~\cite{transformer}. 
	Through the calculations, we obtain four relative position encodings including $\mathbf{p}_{ij}^{\text{tl}}$, $\mathbf{p}_{ij}^{\text{tr}}$, $\mathbf{p}_{ij}^{\text{br}}$, and $\mathbf{p}_{ij}^{\text{bl}}$. The final representation of the relative position bias can be acquired as follow:
	\begin{align}
		& {\mathbf{bb}}_{ij}=\mathbf{W}^{\text{tl}}\mathbf{p}_{ij}^{\text{tl}}+\mathbf{W}^{\text{tr}}\mathbf{p}_{ij}^{\text{tr}}+\mathbf{W}^{\text{br}}\mathbf{p}_{ij}^{\text{br}}+\mathbf{W}^{\text{bl}}\mathbf{p}_{ij}^{\text{bl}} \\
		& {e}_{ij}^{'}={e}_{ij}+\left( \mathbf{W}^{\text{q}}\mathbf{m}_{i}^{l} \right) ^{\top}\mathbf{bb}_{ij}
	\end{align}
	where $\mathbf{W}^{\text{tl}}$, $\mathbf{W}^{\text{tr}}$, $\mathbf{W}^{\text{br}}$ and $\mathbf{W}^{\text{bl}}$ are learnable matrices. ${e}_{ij}^{'}$ is the final attention coefficient. 
	
	In previous works, every node (including both word-level and region-level) can attend to each other, neglecting the document structure information. We inject the graph structure into the attention mechanism by performing masked attention --- we only compute ${e}_{ij}^{'}$ for nodes $j\in \mathcal{N}\left( i \right) $, where $\mathcal{N}\left( i \right)$ is the neighbor area of node $i$ in the document. In our implementation, we select the top-$k$ nodes nearest to node $i$ (including itself) according to the Euclidean distance. Moreover, it is worth noting that we also append a global node $\mathbf{m}_{0}^{l}$ to $\mathcal{N}\left( i \right)$ to assist the model in understanding a document from the global aspect. Finally, the output vectors $\mathbf{h}_{i}^{l}$ are obtained as follow:
	
	\begin{align}
		& \mathbf{\hat{h}}_{i}^{l}=\sum_j{\frac{\exp \left( e_{ij}^{'} \right) \mathbf{m}_{j}^{l}}{\sum_k{\exp \left( e_{ik}^{'} \right)}}\mathbf{W}^{\text{v}}}\quad \quad \quad j,k\in \mathcal{N}\left( i \right) \\
		& \mathbf{h}_{i}^{l}=\text{LN}\left( \mathbf{\hat{h}}_{i}^{l}+\text{FFN}\left( \mathbf{\hat{h}}_{i}^{l} \right) \right) 
	\end{align}
	in which $\mathbf{W}^{\text{v}}\in \mathbb{R}^{d\times d}$, LN is layer normalization, FFN is feed-forward network~\cite{transformer}.

	\subsection{Pre-training task}
	Following Self-Doc~\cite{selfdoc}, we use the Masked Sentence Modeling (MSM) as the pre-training task for the GraphDoc to learn the language representation with the clues of visual embeddings and sentence embeddings. During the pre-training stage, each sentence is randomly and independently masked, while its corresponding layout information is preserved. For the masked sentence, its text content is replaced with a special symbol named [MASK]. The training target is to predict the sentence embeddings of masked ones based on the sentence embeddings and the visual embeddings of others. In this way, the GraphDoc can understand the semantic contexts by fully utilizing the multimodal information. We apply the smooth L1~\cite{fastrcnn} to minimize the pre-training loss as follow:
	\begin{equation}
		\mathcal{L}_{\text{MSM}}\left( \Theta \right) =\sum_i{\text{smooth}_{L_1}\left( \mathbf{s}_i-f_{\text{GraphDoc}}\left( \mathbf{s}_i|\overline{\mathbf{S}},\mathbf{V} \right) \right)}
	\end{equation}
	where $\Theta$ is the trainable parameter set of GraphDoc and $f_{\text{GraphDoc}}\left(\cdot \right)$ outputs the predicted sentence embedding of masked ones, $\overline{\mathbf{S}}$ is the sentence embedding of unmasked ones.

	\begin{figure*}[!htp]
		\centerline{\includegraphics[width=1\linewidth]{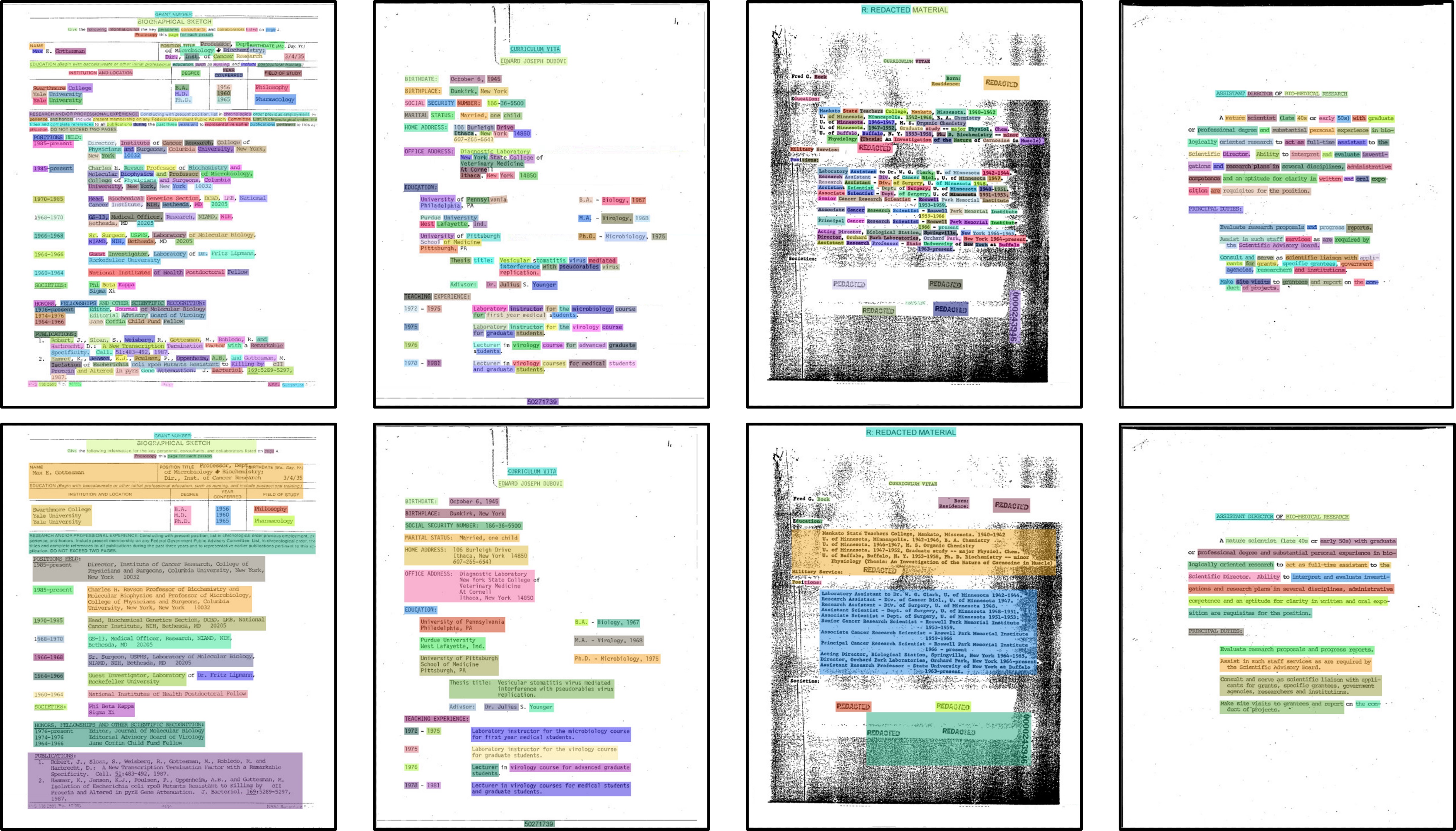}}
		\caption{Some examples of word-level and region-level detection results by EasyOCR in the RVL-CDIP dataset. \textbf{First Row:} The detection results in word-level. \textbf{Second Row:} The detection results in region-level.}
		\label{region_samples}
	\end{figure*}

\section{Experiments}
	\subsection{Datasets}
	We will introduce several datasets that are used for pre-training and evaluating our GraphDoc in this section. The extensive experiments are conduct on four benchmark datasets: RVL-CDIP~\cite{rvlcdip}, FUNSD~\cite{funsd},SROIE~\cite{sroie}, and CORD~\cite{cord}.
	
	\textbf{PubLayNet}
	The PubLayNet dataset~\cite{publaynet} contains over 360k scholarly articles with bounding boxes on 5 categories, such as text block, heading, figure, list, and table. An object detection task is defined on PubLayNet. We use this dataset to pre-train the visual backbone.
	
	\textbf{RVL-CDIP}
	The RVL-CDIP dataset~\cite{rvlcdip} consists of 400k scanned document images, including 320k training images, 40k validation images, and 40k test images. The images are categorized into 16 classes, with 25k images per class. A multi-class single-label classification task is defined on RVL-CDIP.
	
	\textbf{FUNSD}
	The FUNSD~\cite{funsd} is a dataset for form understanding in noisy scanned documents. It consists of 199 real, fully annotated, scanned form images. The dataset is split into 149 training samples and 50 testing samples. It is suitable for various tasks, but we focus on the entity labeling task in this paper.
	
	\textbf{SROIE}
	The SROIE dataset is composed of 626 receipts for training and 347 receipts for testing. Every receipt contains four predefined target fields: company, date, address, and total. The segment-level text bounding box and the corresponding transcript are provided. The task is to label each word to the right field.
	
	\textbf{CORD}
	The CORD dataset contains 800/100/100 receipts for training/validation/testing. The receipts are labeled with 30 types of entities under 4 categories: company, date, address, and total. A list of text lines with bounding boxes is provided. The task is the same as SROIE.

	\subsection{Implementation details}
	We initialize the Sentence-BERT with BERT-NLI-STSb-base~\footnote{\label{sentence-bert-pretrained}\url{https://github.com/UKPLab/sentence-transformers}} pre-trained for NLI~\cite{nli} and STS-B~\cite{sts-b}. A document object detector~\cite{centernet} using the backbone of Swin Transformer~\cite{swintransformer} with FPN is trained on the PubLayNet dataset, which will be used as the visual information extractor of the GraphDoc. We build our pre-training corpus based on the training set of RVL-CDIP~\cite{rvlcdip} with 320k images. The EasyOCR~\cite{easyocr} engine is used to extract the bounding boxes and text contents. There are two types of bounding boxes, word-level and region-level, in EasyOCR. Some results of these two types of bounding boxes are visualized in Figure~\ref{region_samples}. In our experiments, we use the region-level bounding boxes in default. 
	
	During pre-training, we freeze the parameters of Sentence-BERT and jointly train the visual backbone and GraphDoc in an end-to-end fashion. GraphDoc contains 12 layers of graph attention blocks, with the hidden size set to 768 and the number of heads to 12. Moreover, the parameter top-$k$ for our graph attention layer is set to 36. As for the MSM task, following the setting in BERT~\cite{bert}, 15\% of all input sentences are masked among which 80\% are replaced by the [MASK] symbol, 10\% are replaced by random sentences from other documents, and 10\% remain the same. We pre-train the GraphDoc using Adam optimizer~\cite{adam,adamv2}, with the learning rate of $5\times 10^{-5}$. The learning rate is linearly warmed up over the first $10\%$ steps then linearly decayed. The pre-training is conducted on 4 Telsa A100 48GB GPUs with a batch size of 120, and it takes around 10 hours to complete the pre-training for 10 epochs.
	
	\subsection{Ablation study}
	To verify the effectiveness of each component, we conduct ablation experiments through several designed systems as shown in Table~\ref{ablation-study}. The model is not modified except for the component being tested. The model's performance is evaluated on the FUNSD dataset, and the training details will be elaborated in the next subsection.
	
	\begin{table}[htp]
		\centering
		\renewcommand\arraystretch{1.3}
		\setlength{\tabcolsep}{2.1mm}
		
		\caption{Comparison of F1 among systems from T1 to T6 on the FUNSD dataset. Attributes for comparison include: 1) employing the textual encoder; 2) employing the visual encoder; 3) jointly optimizing (JO) the visual encoder 4) using the graph attention layer (GAT); 5) using the relative position encoding (RPE); 6) pre-training the model.}
		\label{ablation-study}
		\begin{tabular}{cccccccc}
			\hline
			\multirow{2}{*}{System} & \multicolumn{2}{c}{Encoder} & \multirow{2}{*}{JO} & \multirow{2}{*}{GAT} & \multirow{2}{*}{RPE}      & \multirow{2}{*}{Pre-train} & \multirow{2}{*}{F1} \\ \cline{2-3}
			& Text & Vision & & & & \\ \hline
			T1 & \checkmark & -          & -		  & \checkmark & \checkmark & \checkmark & 86.36 \\
			T2 & \checkmark & \checkmark & -		  & \checkmark & \checkmark & \checkmark & 86.13 \\
			T3 & \checkmark & \checkmark & \checkmark & -          & -          & \checkmark & 85.33 \\
			T4 & \checkmark & \checkmark & \checkmark & \checkmark & -          & \checkmark & 86.56 \\
			T5 & \checkmark & \checkmark & \checkmark & \checkmark & \checkmark & -          & 80.66 \\
			T6 & \checkmark & \checkmark & \checkmark & \checkmark & \checkmark & \checkmark & \textbf{87.77} \\ \hline
		\end{tabular}
	\end{table}
	
	\textbf{The effectiveness of multimodality}
	To evaluate the effect of multimodality, we design the systems T1 and T6 as shown in Table~\ref{ablation-study}. Each system is designed with or without the visual encoder. When both text and image modalities are encoded by our encoder, the model (T6) exhibits better performance. This illustrates that the multimodal pre-training in GraphDoc learns better interactions from different modalities, thereby leading to better performance.
	
	\textbf{The effectiveness of joint optimization}
	Different from the previous work~\cite{selfdoc}, we jointly optimize the visual encoder with the GraphDoc. To evaluate the effect of joint optimization, we design the systems T2 and T6 as shown in Table~\ref{ablation-study}. When the parameters of visual encoder are not updated with the GraphDoc, the performance drops from 87.77 (T6) to 86.13 (T2).
	 
	\textbf{The effectiveness of gate fusion layer}
	To evaluate the effectiveness of the gate fusion layer, we conduct experiments to compare it with two common feature fusion strategies~\cite{vibertgrid, m2docclassify, m2fcn}, including addition and concatenation as shown in Table~\ref{gate_fusion} . The gate fusion outperforms other strategies by a large margin.
	
	\textbf{The effectiveness of graph attention network}
	To investigate the effect of the proposed graph attention layer (GAT), we designed the systems T3 and T4 as shown in Table~\ref{ablation-study}. Different from the T4, T3 uses the original Transformer architecture~\cite{transformer} instead of the GAT. When applying the GAT, the model (T4) achieves better performance, which demonstrates the effectiveness of the proposed GAT. This is mainly because a text block in a document relies more heavily on its surrounding contexts, and the designed GAT obligates each node to attend to only its neighborhoods. Moreover, through appending a global node and stacking $N$ GAT layers, the model can capture the global information of the document as well.
	
	\begin{table}[t]
		\centering
		\renewcommand\arraystretch{1.5}
		\setlength{\tabcolsep}{3.0mm}
		
		\caption{Performance by using different feature fusion strategies in the system T6 on the FUNSD dataset.}
		\label{gate_fusion}
		\begin{tabular}{|c|c|c|c|}
			\hline
			Method & Addition & Concatenation & Gate Fusion    \\ \hline
			F1     & 86.23    & 83.79         & \textbf{87.77} \\ \hline
		\end{tabular}
	\end{table}

	\textbf{The effectiveness of relative position encoding}
	To investigate the effect of relative position encoding (RPE), we designed the systems T4 and T6 as shown in Table~\ref{ablation-study}. When the RPE is used, the model (T6) achieves better performance. This is mainly because the RPE encodes the relative positions between bounding boxes into attention scores, which will further boost the model's awareness of the relationship between nodes.
	
	\textbf{The effectiveness of pre-training}
	Since the Sentence-BERT and the visual backbone in GraphDoc have been pre-trained from a large corpus, it's doubtful whether GraphDoc needs extra pre-training. As shown in Table~\ref{ablation-study}, we design T5 and T6 to answer this question. T6 outperforms T5 by a large margin, which demonstrates the necessity of pre-training in GraphDoc. It is worth noting that two layout embedding layers, gate fusion layers across GAT layers in GraphDoc are initialized randomly. The model needs pre-training to learn a generic representation on layout embedding layers and make gate fusion layers with GAT more capable of multimodal interaction.

	\textbf{The effectiveness of residual connection}
	We believe that visual dependencies might be different across $N$ GAT layers. To verify this assumption, we design 5 systems with residual connections across different numbers of GAT layers as shown in Table~\ref{residual_connection}. As the number of residual connection layers increases, the performance of the model (T6) becomes better, which demonstrates the effectiveness of the residual connection.

	\textbf{The impact of Top-$K$}
	To investigate the effect of the configuration top-$k$ in graph attention layer, as shown in Table~\ref{topk},  we set a different number of top-$k$ in the T6 system and evaluate on the FUNSD dataset. When top-$k$ is too small, the performance of the model degrades due to the limited receptive field of each node. When the top-$k$ increases, especially when the top-$k$=256, it is essentially the system T3 with RPE, which further illustrates the effectiveness of the proposed GAT.

	\begin{table}[t]
		\centering
		\renewcommand\arraystretch{1.5}
		\setlength{\tabcolsep}{3.0mm}
		
		\caption{Performance by varying number of residual connection in the system T6 on the FUNSD dataset.}
		\label{residual_connection}
		\begin{tabular}{|c|c|c|c|c|c|}
			\hline
			Num & 1    & 3    & 6    & 9    & 12    \\ \hline
			F1  & 86.44 & 86.83 & 87.14 & 87.32 & \textbf{87.77} \\ \hline
		\end{tabular}
	\end{table}
	
	\begin{table}[t]
		\centering
		\renewcommand\arraystretch{1.5}
		\setlength{\tabcolsep}{2.6mm}
		
		\caption{Performance by varying the top-$K$ in the system T6 on the FUNSD dataset.}
		\label{topk}
		\begin{tabular}{|c|c|c|c|c|c|c|}
			\hline
			Top-$K$ & 16    & 24    & 36    & 48    & 60 & 256    \\ \hline
			F1   & 85.68 & 86.58 & \textbf{87.77} & 87.26 & 87.12 & 86.87 \\ \hline
		\end{tabular}
	\end{table}

	\subsection{Comparison with state-of-the-art methods}
	We compare our method with other state-of-the-art methods on three document understanding tasks, such as Form Understanding, Receipt Understanding and Document Classification. The results are shown in Table~\ref{comparsion}. In order to form a fair comparison, we also present the results of GraphDoc using ResNet-50 as the visual backbone, as shown in ``GraphDoc$_{\text{ResNet}}$'' in Table~\ref{comparsion}.
	
\begin{figure}[t]
		\centerline{\includegraphics[width=0.9\linewidth]{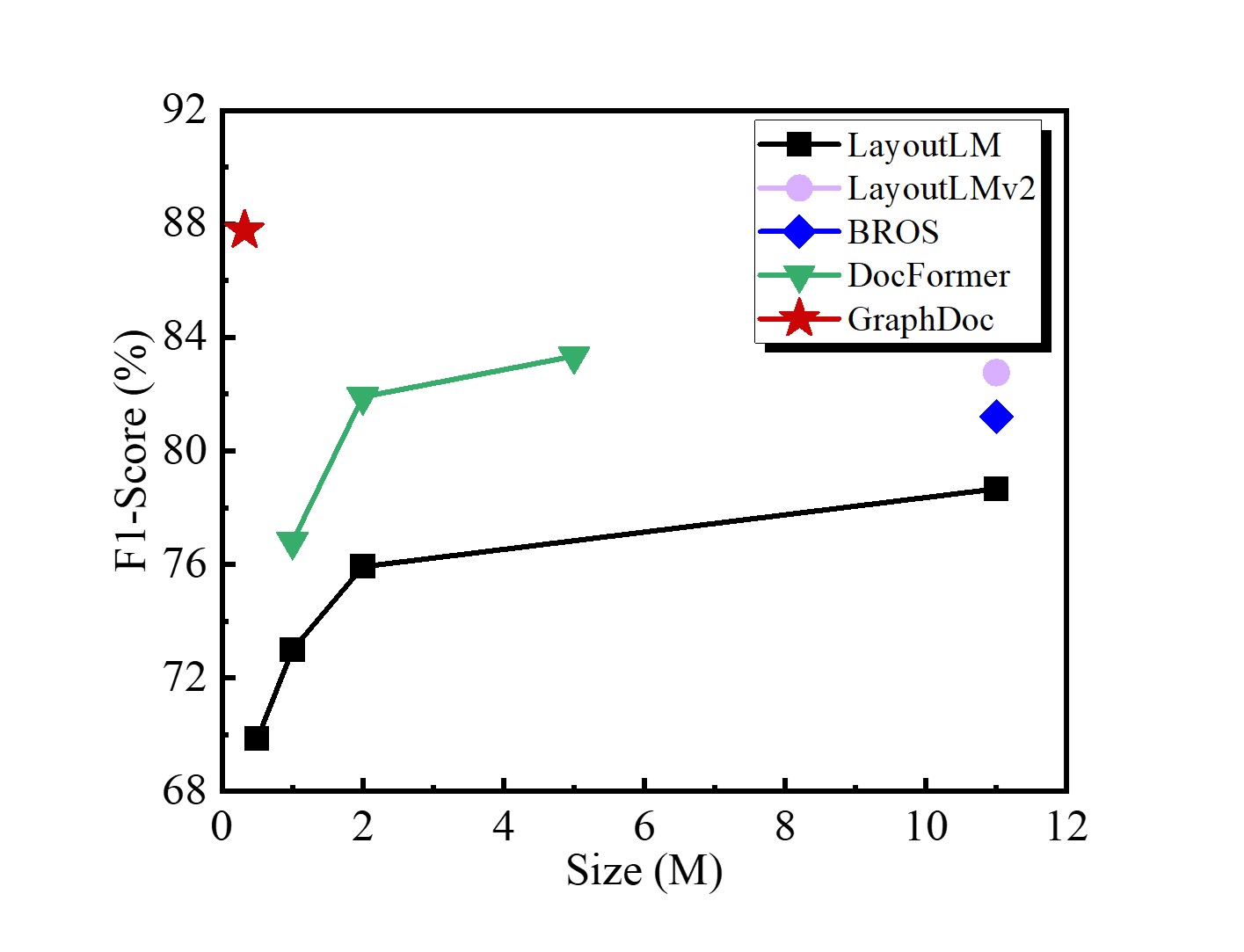}}
		\caption{Performance by varying size of pre-train data on the FUNSD dataset.}
		\label{pretrain_data}
	\end{figure}

	\textbf{Form Understanding} Form understanding requires the model to predict the label for each semantic entity. We use FUNSD~\cite{funsd} as the evaluation dataset. The officially-provided OCR texts and bounding boxes are used during training and testing. We take the semantic entities as input and feed the final output representations of GraphDoc to a classifier. We apply cross-entropy loss for finetuning. The model is finetuned for 50 epochs with a learning rate of $5\times 10^{-5}$ and the batch size of 2. All the parameters except Sentence-BERT are trained. We use entity-level F1 score as the evaluation metric. Table~\ref{comparsion} lists the entity-level F1 score on the FUNSD. It is worth noting that the Text+Layout+Image models outperform both Text and Text+Layout models generally, which demonstrates the indispensability of multimodal modeling in document understanding. Moreover, under the same modality setting (Text+Layout+Image), GraphDoc also outperforms existing multimodal approaches and achieves the new state-of-the-art result, which demonstrates the effectiveness of the proposed model. The systems (LayoutLM, LayoutLMv2, BROS, etc.) designed in token-level are pre-trained on a large corpus (11M), which increases a lot of pre-training time. For example, LayoutLMv2 takes about 500 hours to complete pre-training on 4 Telsa A100 48GB GPUs, while GraphDoc only needs 10 hours. When the size of pre-train data is not sufficient, the performance of token-level systems decreases significantly as shown in Figure~\ref{pretrain_data}.
	
	\begin{table*}[htp]
		\centering
		\renewcommand\arraystretch{1.5}
		\setlength{\tabcolsep}{3mm}
		
		\caption{Comparison with state-of-the-art methods. \textbf{Bold} indicates the SOTA and \underline{underline} indicates the second best.}
		\label{comparsion}
		\begin{tabular}{l|c|c|c|c|c|c|c|c}
			\hline
			\multirow{2}{*}{Model} & \multirow{2}{*}{Modality} & \multirow{2}{*}{\#Data} & \multirow{2}{*}{Scale} & FUNSD & SROIE & CORD  & RVL-CDIP & \multirow{2}{*}{\#Params} \\ \cline{5-8} & & & & F1 & F1 & F1 & Accuracy &                           \\ \hline
			BERT~\cite{bert}                  & \multirow{2}{*}{Text} & - & Token & 60.26 & 90.99 & 89.68 & 89.81 & 110M \\
			RoBERTa ~\cite{roberta}           &                       & - & Token & 66.48 & -     & -     & 90.06 & 125M \\ \hline
			LayoutLM~\cite{layoutlm}  & \multirow{5}{*}{Text+Layout}  & 11M   & Token & 78.66 & 94.38 & 94.72 & 94.42 & 113M \\
			BROS~\cite{bros}                 &                        & 11M   & Token & 81.21 & 95.48 & 95.36 & 95.58 & 139M \\
			StructureLM~\cite{structurallm}  &                        & 11M   & Token & 85.14 & -     & -     & 96.08 & 355M \\ 
			LiLT~\cite{wang2022lilt}         &                        & 11M   & Token & \textbf{88.41}& -     & 96.07 & 95.68 & -  \\ 
			FormNet~\cite{lee_formnet_2022}  &                        & 700K  & Token & 84.69 & -     & \textbf{97.10}& - & 217M  \\ 
			\hline
			LayoutLMv2~\cite{layoutlmv2}     & \multirow{4}{*}{Text+Layout+Image} & 11M & Token & 82.76 & 96.25 & 94.95 & 95.25 & 200M \\
			DocFormer~\cite{docformer}       &                        & 5M  & Token & 83.34 & -     & 96.33 & \textbf{96.17 }& 183M \\
			Self-Doc~\cite{selfdoc}          &                        & 320K  & Region& 83.36 & -     & -     & 92.81 & -    \\
			UniDoc~\cite{unidoc}             &                        & 300K  & Region& 87.38 & -     & 96.64 & 93.92 & 274M  \\ 
			\hline
			GraphDoc$_{\text{ResNet}}$       & \multirow{2}{*}{Text+Layout+Image}     & 320K  & Region& \underline{87.95} & \underline{98.41}     & 96.56 & \underline{96.10}     & 262M     \\
			GraphDoc                         &                        & 320K  & Region& 87.77 & \textbf{98.45} & \underline{96.93} & 96.02     & 265M     \\ \hline
		\end{tabular}
	\end{table*}

	\textbf{Receipt Understanding} Receipt understanding requires the model to recognize a list of text lines with bounding boxes. The performance of this task is evaluated on SROIE~\cite{sroie} and CORD~\cite{cord} datasets. Like FUNSD, we use officially-provided OCR annotations and bounding boxes for fine-tuning and feed the output representations of GraphDoc to the classifier. The model is finetuned for 50 epochs with a batch size of 4 and a learning rate of $5\times 10^{-5}$. The evaluation metric is the entity-level F1 score. Table~\ref{comparsion} shows the model accuracy on both SROIE and CORD datasets. Our model achieves the new state-of-the-art results on the SROIE dataset in the existing works of literature. We also achieve second place in the public leader board in Task-3 on SROIE just by a single mdoel~\footnote{\label{sroie_borad}\url{https://rrc.cvc.uab.es/?ch=13&com=evaluation&task=3}}.
	
	\textbf{Document Classification} Document classification involves predicting the category for each document image. We use RVL-CDIP~\cite{rvlcdip} as the target dataset. The OCR words and bounding boxes are extracted by EacyOCR~\cite{easyocr}. We feed the global node of output representations of GraphDoc to the classifier. We fine-tune the model for 30 epochs with a batch size of 64 and a learning rate of $1\times 10^{-5}$. Classification accuracy over 16 categories is used to measure model performance. Table~\ref{comparsion} shows the model accuracy on RVL-CDIP datasets and GraphDoc achieves a state-of-the-art result. The reason why the performance of Self-Doc is worse than other Text+Layout models is mainly Self-Doc uses the fixed visual encoder without learning a suitable representation in vision modality for downstream tasks. While in GraphDoc, we jointly train our visual backbone. It is worth noting that UniDoc~\cite{unidoc} does not have the [CLS] token for classification, and it simply uses the overall representation by averaging all output region features and learns a classifier on top of the overall representation with cross-entropy loss. In this way, it implicitly agrees that each region is equally important for the document classification, which is the main reason for its poor performance compared to other Text+Layout+Image systems on the RVL-CDIP dataset.
	
	\begin{figure*}[t]
		\centerline{\includegraphics[width=1\linewidth]{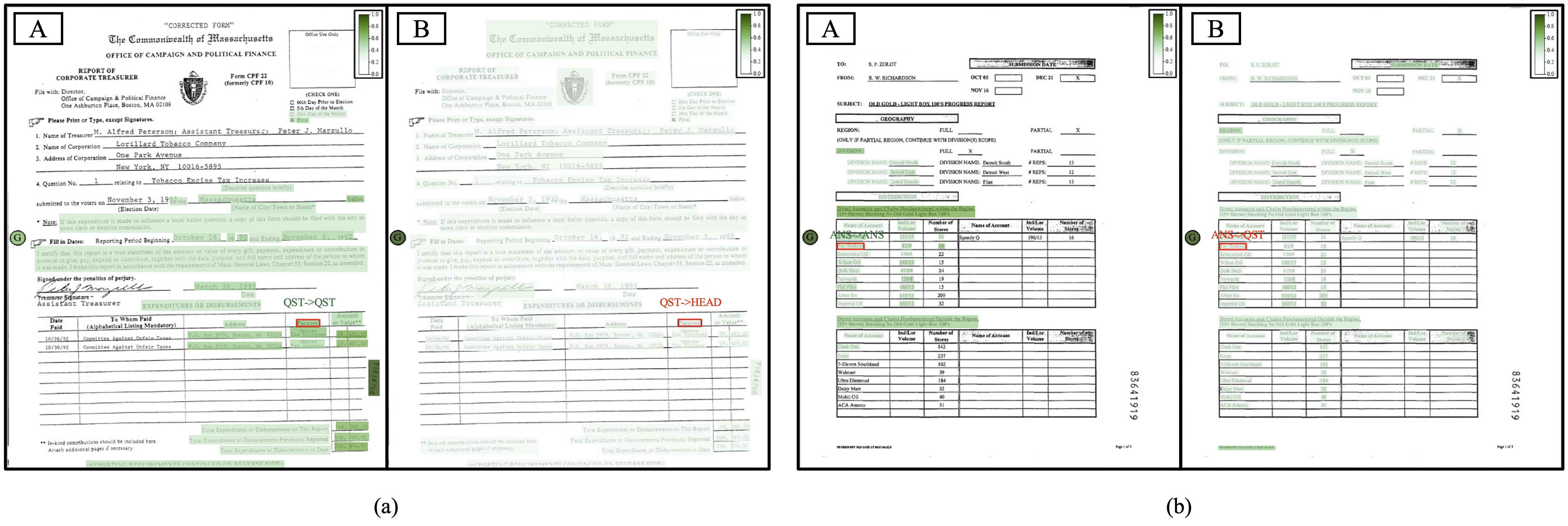}}
		\caption{Attention visualization results on the FUNSD dataset. The A/B series refers to the visualization of GAT/Transformer, respectively. The gradation of color of each text node indicates varied attention weight from the chosen text node (emphasized with a red bounding box). The global node is attached to the left side of each document image. 
			Maps from the ground-truth label to the system predicted label are annotated near the target area.
			The green text indicates correct predictions, while red indicates incorrect predictions.
		}
		\label{vis_atten}
	\end{figure*}

	\subsection{Case study}
	
	\subsubsection{GAT Vs. Transformer}
	The GAT is designed to force each text node in the document into attending more accurately on neighborhood area. As shown in Figure~\ref{vis_atten}, we give some attention visualization results of Transformer and GAT on the same text node of certain documents on the FUNSD dataset. Each visualization result is obtained using the averaged attention weights from the last attention layer of each model. From the attention results listed in Figure~\ref{vis_atten}, we can find that the Transformer model tends to rely more on the global information and attends homogeneously to each text node in the document, while the GAT model tends to focus on those text nodes which are most relevant of the chosen text node. Specifically, as shown in Figure~\ref{vis_atten}(a), the GAT model attends mostly to the surrounding area, including the contents of the table and the corresponding values of the chosen text node ``Purpose'' and classifies it rightly into class ``Question'', while the Transformer model predicts it as class ``Header'' since it attends to too many useless text nodes. Similar situations can be observed in Figure~\ref{vis_atten}(b).

	\subsubsection{Region Vs. Word}
\begin{figure*}[htp]
		\centerline{\includegraphics[width=1\linewidth]{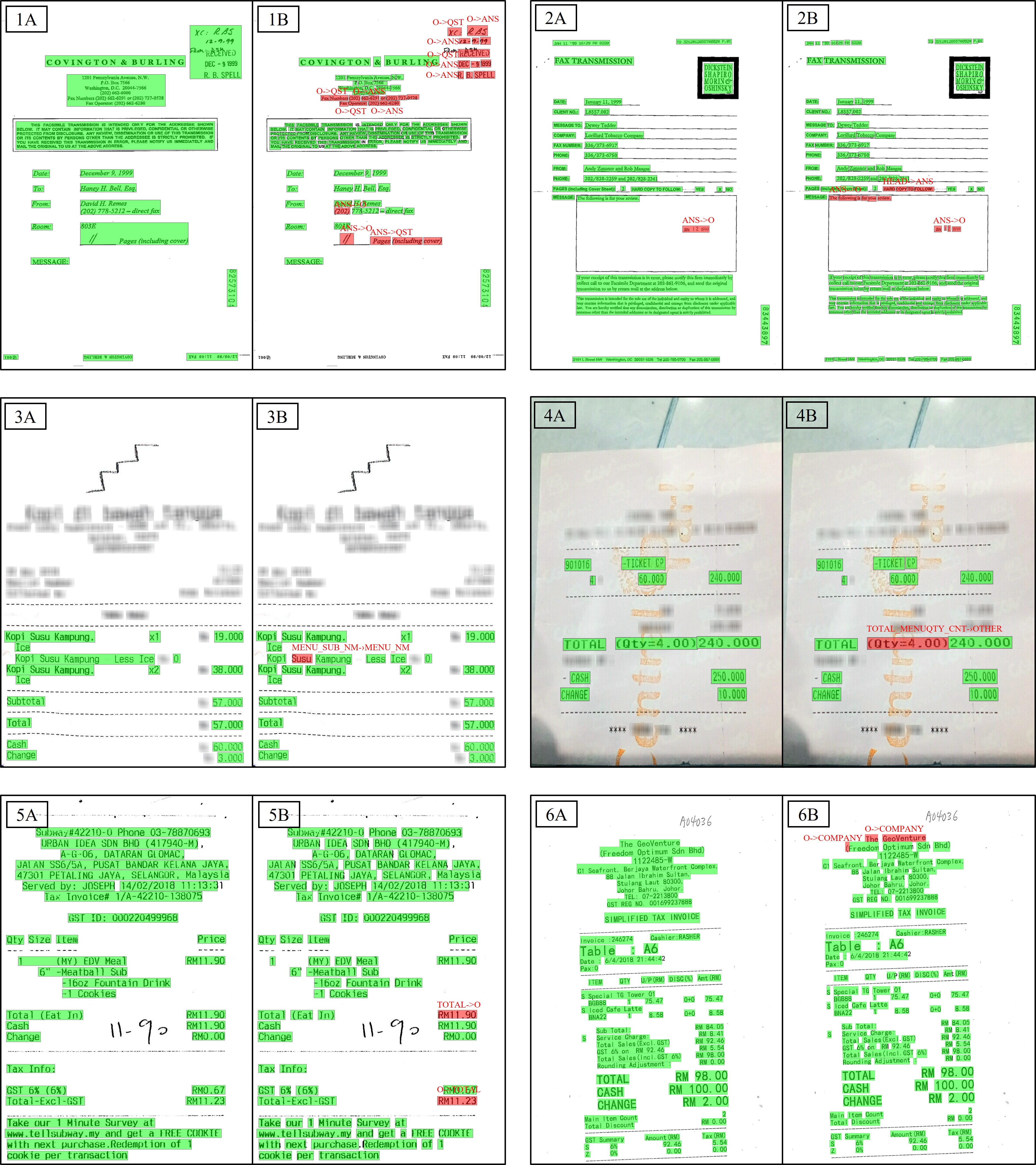}}
		\caption{Visualization results on downstream tasks. Samples 1-2/3-4/5-6 are from the FUNSD/CORD/SROIE dataset, and the A/B series refers to the visualization of GraphDoc/LayoutLMv2 systems, respectively. The green shaded area in each image represents the correct classification results, while the red parts are predicted wrongly. Maps from the ground-truth label to the system predicted label are annotated near the wrongly classified area. Best viewed in color.}
		\label{case_study}
	\end{figure*}

	One important motivation behind GraphDoc is to explore the advantage of region-level modeling versus word-level modeling across scanned document images. As mentioned above, our method has achieved new state-of-the-art performance on several downstream tasks, surpassing word-level modeling methods such as LayouLM by a large margin. We visualize several document samples from three different downstream tasks in Figure~\ref{case_study} to verify this. As the top-left images 1A and 1B depict, GraphDoc classifies semantic entities in region-level and correctly predicts ``803E // Pages (including cover)" to \textit{Answer} category by utilizing the prior knowledge that these words are in the same semantic region. However, without paying special attention to region-level information, LayoutLM-V2 missed ``// Pages (including cover)". Similar situations can be observed in other visualized cases in Figure~\ref{case_study}. Moreover, when both the region-level and word-level boxes are the same as shown in samples 4-5 in Figure~\ref{case_study}, GraphDoc still performs better than LayoutLMv2. 
	
	\subsubsection{Representative Failure Cases}
\begin{figure*}[htp]
		\centerline{\includegraphics[width=1\linewidth]{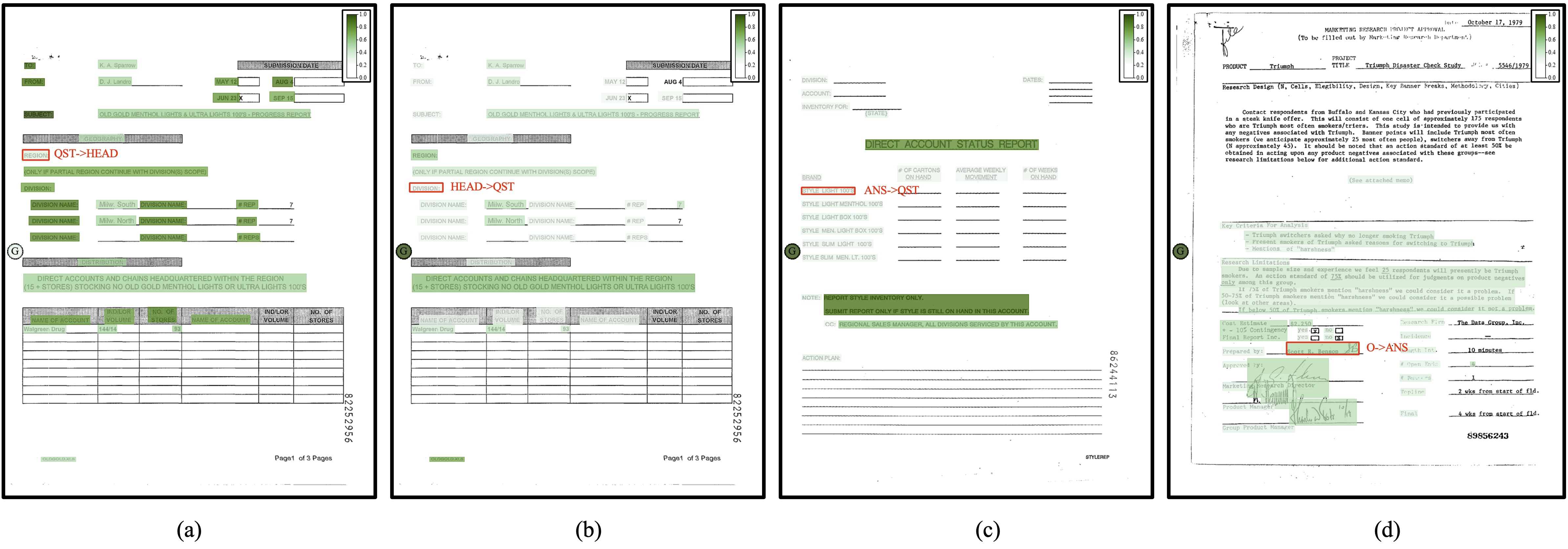}}
		\caption{
			Some representative failure cases of GraphDoc model on the FUNSD dataset. The gradation of color of each text node indicates varied attention weight from the chosen text node (emphasized with a red bounding box). Maps from the ground-truth label to the system predicted label are annotated near the wrongly classified area. Best viewed in color.
		}
		\label{bad_case}
	\end{figure*}

	We have listed some representative failure cases in the FUNSD dataset as shown in Figure \ref{bad_case}. 
	The first kind of failure cases is caused by the structure nesting problem.
	As Figure \ref{bad_case}(a) and \ref{bad_case}(b) depict, the words `REGION' and `DIVISION' are subtitles of `GEOGRAPHY', which should play the same semantic role in the document. However, they are labelled differently as `Question' and `Header' entity. This kind of labelling fuzziness results in some failure cases in GraphDoc such as predicting `Question' entity as `Header' entity in Figure \ref{bad_case}(a). 
	The second kind appears in table-like document as shown in Figure \ref{bad_case}(c). It's not easy to distinguish whether the first row or the first column of one table serves as the key area. GraphDoc also makes a mistake for predicting the second cell of the first column, which is the `Answer' to its upper cell, as the `Question' entity.
	The last kind is caused by some mistakes in ground-truth labelling. In Figure \ref{bad_case}(d), we can see that the name `Scottt R. Benson' is wrongly labelled as an `Other' entity but our GraphDoc model predicts it as the right label `Answer'.
	
\section{Conclusion}
	In this work, we present the GraphDoc, a multimodal graph attention-based model for various Document Understanding tasks. GraphDoc fully utilizes the text, image, and layout information in a document. Considering a text block relies more heavily on its surrounding context, we present a novel graph attention network instead of the Transformer architecture. Each input node can attend to only its neighborhood nodes and a global node, which makes the model learn contextualized information in the document from both local and global aspects. Moreover, we also propose a gate fusion layer for each input node to fuse the textual and visual features. GraphDoc learns a generic representation from only 320k unlabeled documents via the Masked Sentence Modeling task. Extensive experiment results on some document understanding tasks, such as form understanding, receipt understanding, and document classification, show that GraphDoc achieves state-of-the-art, which demonstrates the effectiveness of our proposed method.

\bibliographystyle{IEEEtran}
\bibliography{reference}


\begin{IEEEbiography}[{\includegraphics[width=1.0in,height=1.25in, clip,keepaspectratio]{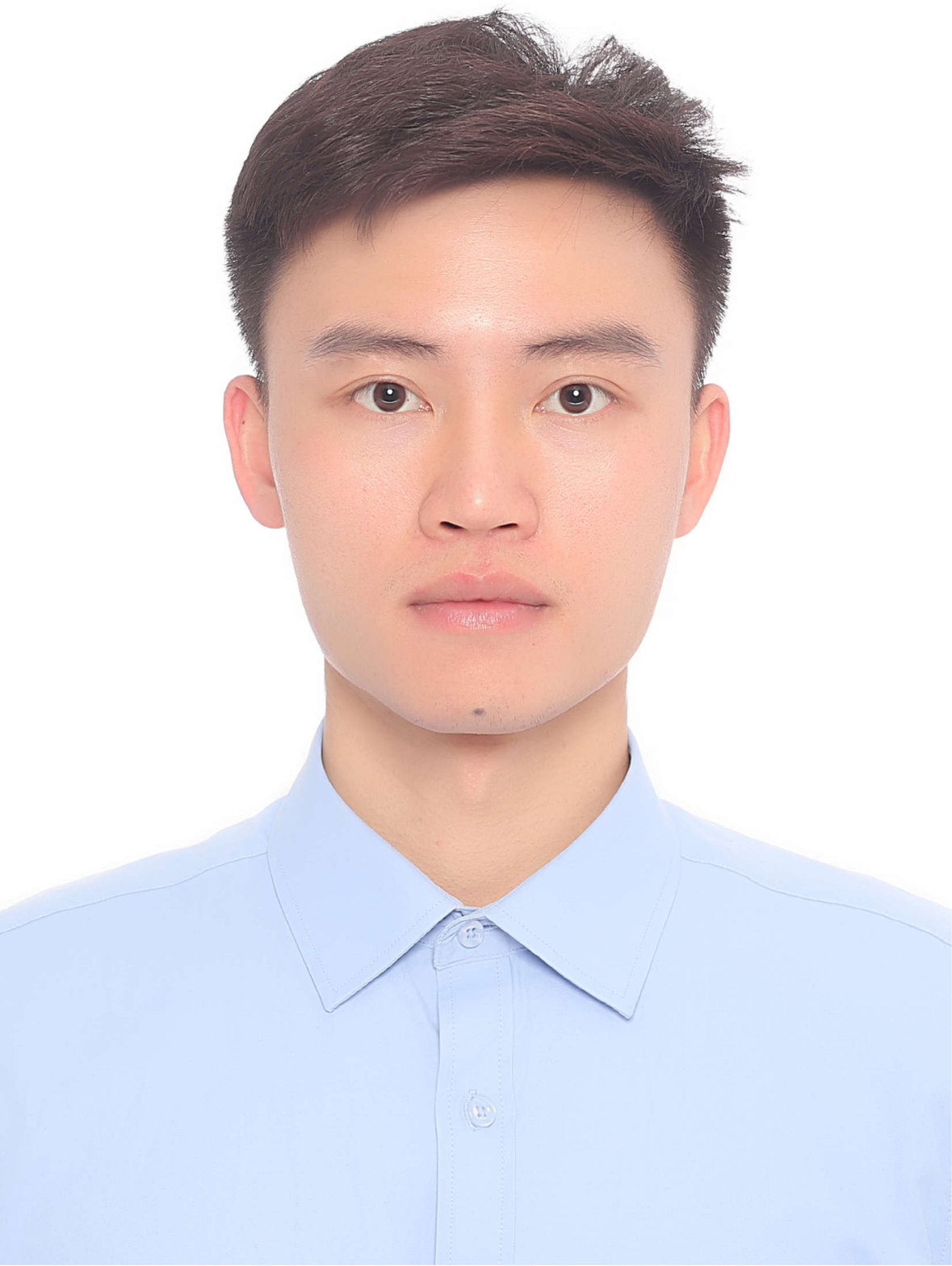}}]{Zhenrong Zhang} received his B.Eng. degree from the Department of Computer Science and Engineering, Northeastern University of China, in 2020. He is currently a Master’s candidate at the University of Science and Technology of China (USTC). His current research area includes document analysis and OCR.
\end{IEEEbiography}

\begin{IEEEbiography}[{\includegraphics[width=1.0in,height=1.25in, clip,keepaspectratio]{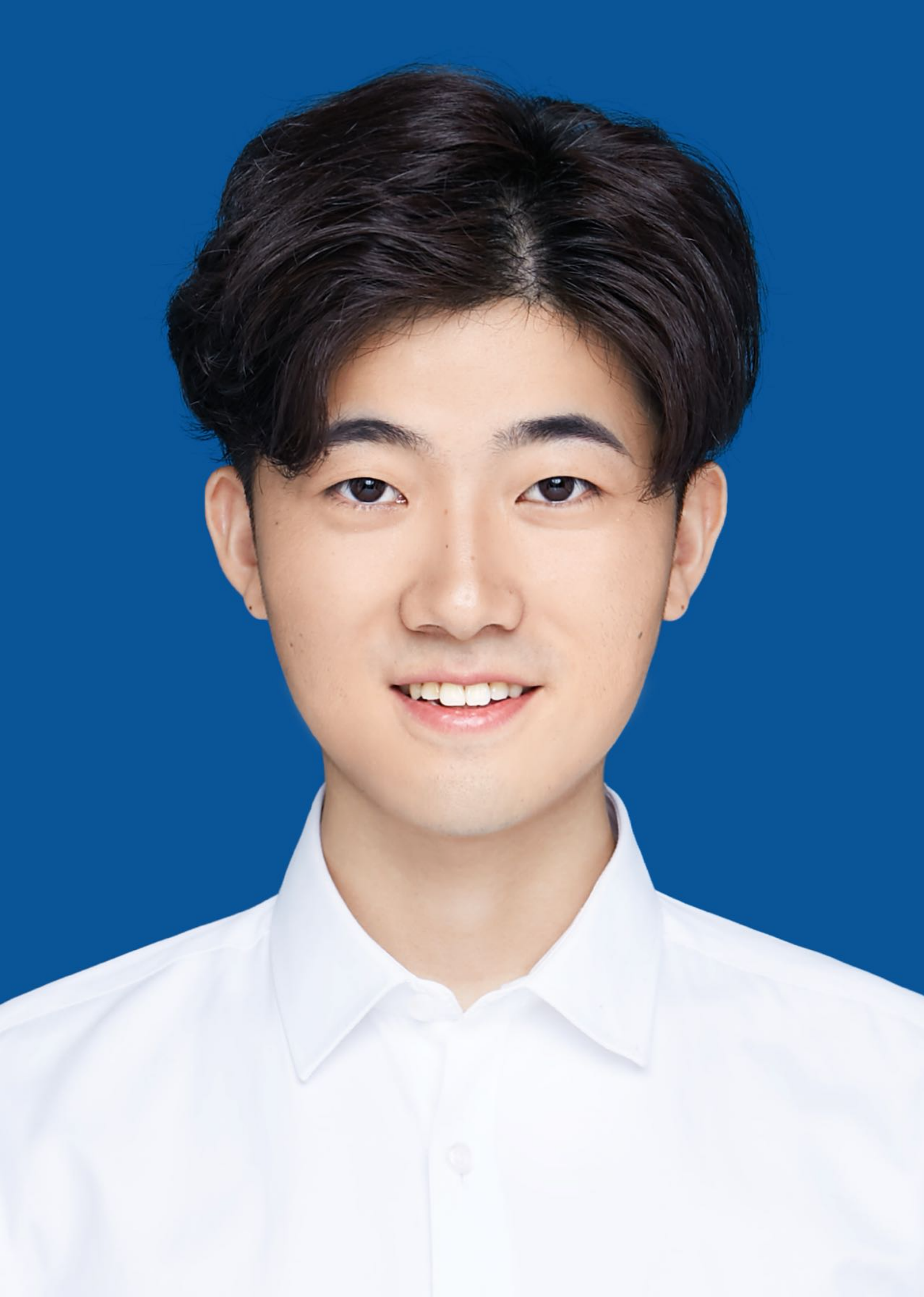}}]{Jiefeng Ma} received his B.Eng. degree from the Department of Electronic Engineering and Information Science, University of Science and Technology of China (USTC) in 2020. He is currently a Master’s candidate at the University of Science and Technology of China (USTC). His current research area includes natural language generation, information extraction, and document analysis.
\end{IEEEbiography}
\begin{IEEEbiography}[{\includegraphics[width=1.0in,height=1.25in, clip,keepaspectratio]{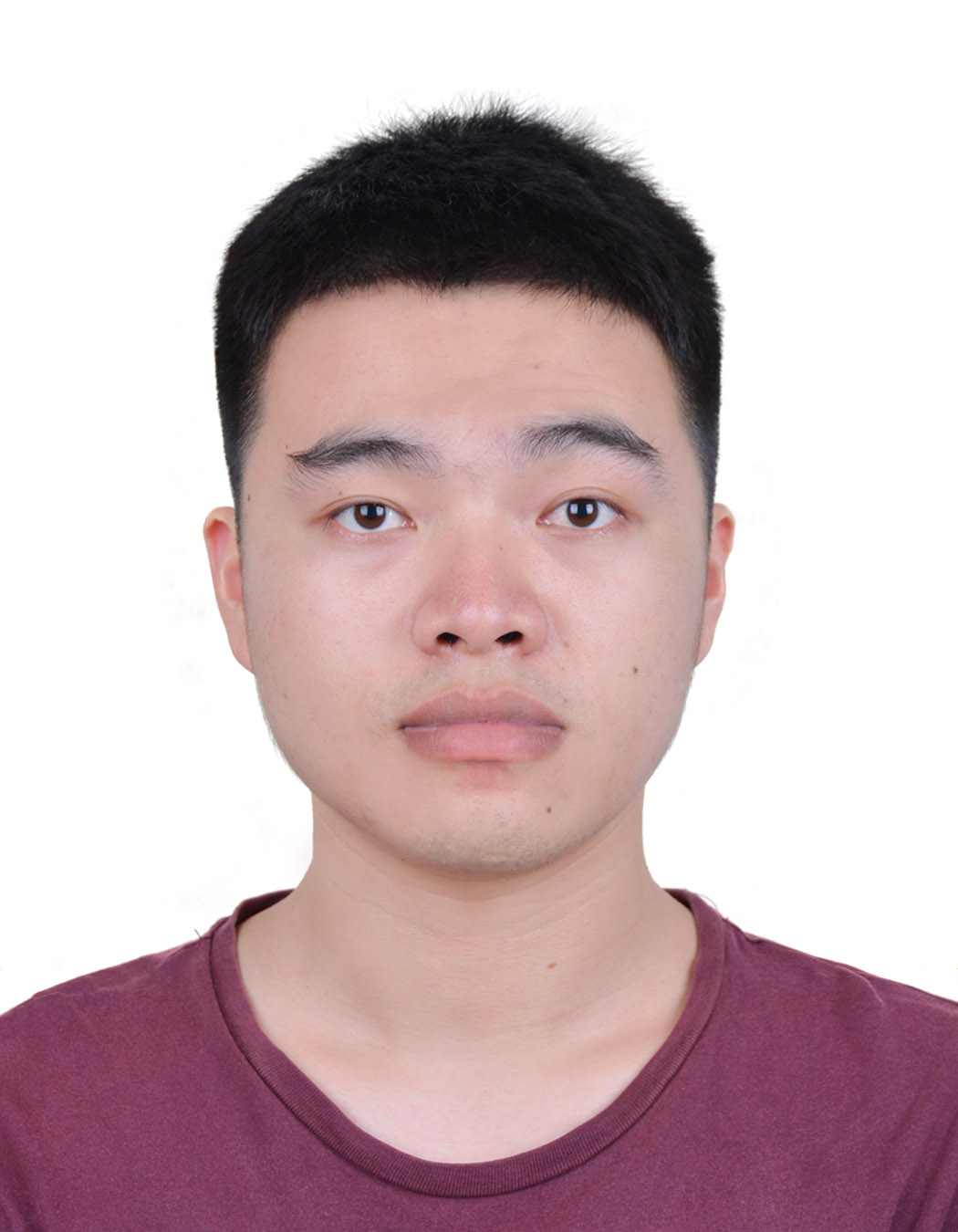}}]
{Jianshu Zhang} received the B.Eng. degree from the Department of Electronic Engineering and Information Science, University of Science and Technology of China (USTC) in 2015. He is currently a Ph.D. candidate of USTC. His current research area is neural network, handwriting mathematical expression recognition and Chinese document analysis.
\end{IEEEbiography}
\begin{IEEEbiography}[{\includegraphics[width=1.0in,height=1.25in, clip,keepaspectratio]{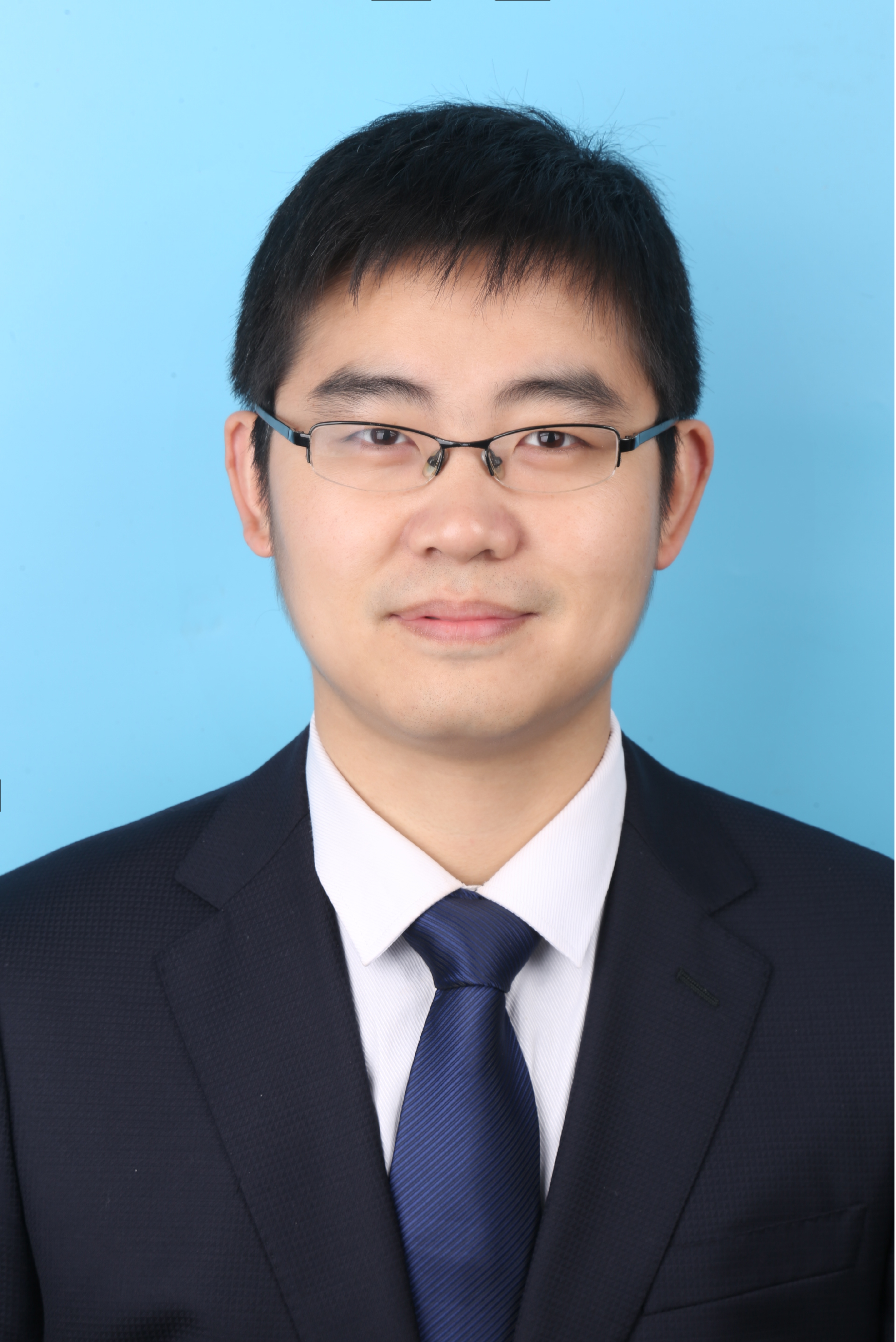}}]{Jun Du} received the B.Eng. and Ph.D. degrees from the Department of Electronic Engineering and Information Science, University of Science and Technology of China (USTC) in 2004 and 2009, respectively. From 2004 to 2009, he was with iFlytek Speech Lab of USTC. During the above period, he worked as an Intern twice for 9 months at Microsoft Research Asia (MSRA), Beijing. In 2007, he also worked as a Research Assistant for 6 months in the Department of Computer Science, The University of Hong Kong. From July 2009 to June 2010, he worked at iFlytek Research on speech recognition. From July 2010 to January 2013, he joined MSRA as an Associate Researcher, working on handwriting recognition, OCR, and speech recognition. Since February 2013, he has been with the National Engineering Laboratory for Speech and Language Information Processing (NEL-SLIP) of USTC.
\end{IEEEbiography}

\end{document}